\documentclass[letterpaper, 10 pt, conference]{ieeeconf}  

\IEEEoverridecommandlockouts                              

\usepackage{amssymb}  
\usepackage{flushend}
\usepackage{amsmath}
\usepackage{graphicx}
\usepackage{multirow}
\usepackage{xcolor}
\usepackage{caption}
\usepackage{subcaption}
\usepackage{cite}
\usepackage{float}

\usepackage[absolute,showboxes]{textpos}

\TPMargin*{3pt}
\newcommand\copyrighttext{
    \footnotesize
    \noindent
    SUBMITTED TO REVIEW AND POSSIBLE PUBLICATION. COPYRIGHT WILL BE TRANSFERRED WITHOUT NOTICE.\\
    Personal use of this material is permitted.
    Permission must be obtained for all other uses, in any current or future media, including reprinting/republishing this material for advertising or promotional purposes, creating new collective works, for resale or redistribution to servers or lists, or reuse of any copyrighted component of this work in other works.}%
\newcommand\copyrightnotice{%
    \begin{textblock*}{6.6in}(0.95in,0.15in)
        \centering
        \copyrighttext
    \end{textblock*}
}

\title{\LARGE \bf
Towards Consistent and Explainable Motion Prediction using Heterogeneous Graph Attention
}

\author{Tobias Demmler$^{1}$, Andreas Tamke$^{2}$, Thao Dang$^{3}$, Karsten Haug$^{4}$ and Lars Mikelsons$^{5}$
\thanks{$^{1}$Tobias Demmler is with Robert Bosch GmbH, Stuttgart, Germany
        {\tt\small tobias.demmler@de.bosch.com}}%
\thanks{$^{2}$Andreas Tamke is with Robert Bosch GmbH, Stuttgart, Germany
        {\tt\small andreas.tamke@de.bosch.com}}%
\thanks{$^{3}$Thao Dang is with the Institute for Intelligent Systems, Department of Computer Science and Engineering, Esslingen University, Germany
        {\tt\small thao.dang@hs-esslingen.de}}%
\thanks{$^{4}$Karsten Haug is with Robert Bosch GmbH, Stuttgart, Germany
        {\tt\small karsten.haug@de.bosch.com}}%
\thanks{$^{5}$Lars Mikelsons leads the Chair of Mechatronics of the University of Augsburg, Germany
        {\tt\small lars.mikelsons@uni-a.de}}%
}

\begin{document}
\copyrightnotice

\maketitle
\thispagestyle{empty}
\pagestyle{empty}

\begin{abstract}

In autonomous driving, accurately interpreting the movements of other road users and leveraging this knowledge to forecast future trajectories is crucial. This is typically achieved through the integration of map data and tracked trajectories of various agents. Numerous methodologies combine this information into a singular embedding for each agent, which is then utilized to predict future behavior. However, these approaches have a notable drawback in that they may lose exact location information during the encoding process. The encoding still includes general map information. However, the generation of valid and consistent trajectories is not guaranteed. This can cause the predicted trajectories to stray from the actual lanes. This paper introduces a new refinement module designed to project the predicted trajectories back onto the actual map, rectifying these discrepancies and leading towards more consistent predictions. This versatile module can be readily incorporated into a wide range of architectures. Additionally, we propose a novel scene encoder that handles all relations between agents and their environment in a single unified heterogeneous graph attention network. By analyzing the attention values on the different edges in this graph, we can gain unique insights into the neural network's inner workings leading towards a more explainable prediction. 

\end{abstract}
\section{INTRODUCTION}
Motion forecasting is an essential part of modern autonomous driving systems and has been experiencing an increase in attention over the past few years. It is crucial for such systems to have a deep understanding of their surroundings in order to make safe decisions. There are two important information sources for motion forecasting. The first is the prior static knowledge of the environment. This is typically captured in a high-definition map (HD map), which contains detailed positional and semantic information about individual lanes of the road. The second one is the dynamic state of all the other traffic participants. This includes the type of the actor (e.g., car, pedestrian, cyclist) and their position and velocity over multiple observed timesteps. Motion forecasting requires predicting the future trajectories of other traffic participants by combining these two information sources. This is challenging due to the multimodal nature of this task and the complexity of road traffic.

Many existing motion prediction models have an architecture with a completely distinct encoder and decoder part \cite{lanegcn}. They encode the entire scene into a single feature vector for each agent and then make their predictions solely based on this feature vector. However, this approach, which condenses the complex scene into a small latent space, often leads to the loss of vital details. For instance, the precise curvature of a relevant lane might be overlooked, causing the prediction to drift into other lanes or off the road. To address this, we propose a refinement module that projects these flawed trajectories back onto the lane graph and employs a Graph Neural Network (GNN) to adjust the trajectories toward a more consistent path based on actual environmental details.

Another challenge with numerous complex motion prediction models is understanding the flow of information within the network, particularly in GNNs with several stages and modules \cite{lanegcn}. We propose a solution to this problem by unifying our approach into a single heterogeneous graph attention network. This not only reduces the complexity of the network but also allows for a straightforward analysis of the information flow among different nodes in the graph by monitoring the attention values. This leads to a more explainable motion forecast.

Thus, our contributions in this paper are twofold:
\begin{itemize}
    \item Firstly, we propose a refinement module that can be readily adapted to improve many existing approaches. This module offers a more accurate and consistent prediction by considering the minute details often lost in traditional approaches. When using this module, it is possible to apply it to an already pretrained network that does not use the refinement module. It is also possible to train the entire network with the refinement module in an end-to-end fashion.

    \item Secondly, we introduce a novel scene encoder that can handle all tasks within a single heterogeneous graph attention network. This approach provides a unique insight into the complex information flow within the GNN by analyzing the attention values, thereby offering a more explainable and transparent model.
\end{itemize} 
\section{RELATED WORK}

Related works can be categorized into multiple groups based on their encoder and decoder architecture.

\subsection{Encoder}

The input data for motion forecasting is rather complex. We usually have environment data represented by an HD map and observed trajectories. The environment is the most challenging part to process since it is represented as a graph without a fixed size. Many neural networks can not handle such data. Multiple approaches have been used over time. In the beginning Convolutional Neural Networks (CNNs) have been used \cite{multipath, Gilles2021HOMEHO, Salzmann2020TrajectronDT}. Later architectures focused on GNN architectures \cite{lanegcn, gilles2022gohome, Cui2022GoRelaGR, gilles2021thomas}. Currently, Transformers are on the rise \cite{shi2023mtr, zhou2023query}, which have many similarities with GNNs. 

The encoding of the observed trajectories is more straightforward since it has a fixed size. It can be done with basic Multi-layer Perceptrons (MLPs) or CNNs with convolutions over time \cite{lanegcn}. Some approaches also combine the trajectory and map data before encoding them together \cite{multipath}.

\subsubsection{Convolutional Neural Networks}

CNNs have been leveraged in earlier works \cite{multipath, Gilles2021HOMEHO, Salzmann2020TrajectronDT}. They have the advantage that they are heavily researched in the computer vision domain \cite{Krizhevsky2012ImageNetCW, resnet}. How they work is also rather intuitive. The environment is rendered as an image in a top-down view. In doing so, the variable-sized graph is transformed into a fixed-sized representation. This image can also include the observed trajectories over time. This approach is no longer used in state-of-the-art approaches for motion forecasting. Many details of the environment get lost in such a rasterized representation, especially if the resolution of the image is on the lower side. Additionally, CNNs are rather expensive to run. First, the semantic graph representation is transformed into an image and then the CNN has to understand the image to generate a semantic understanding. Additionally, most parts of the image are unimportant since they do not contain driveable areas. Therefore, many operations in the CNN do not contribute to understanding the scene.

\subsubsection{Graph Neural Networks}

GNN-based approaches have many advantages over CNN-based ones. Most importantly, they do not discard the semantic information of the scene graph by transforming the graph into an image. Furthermore, they can represent much more details because they are not limited by the resolution of an image. This makes GNNs an ideal approach to encode such structures without losing much information. They are also much more efficient since they do not have to relearn the semantic relations. Additionally, GNN operations are only executed for relevant parts of the scene, whereas many CNN operations are performed on non-driveable areas. A downside of this graph-based approach is that these graph structures can get quite complex. GNNs are also far less researched than CNNs. Despite these drawbacks, graph-based representations allow great flexibility and enable the encoding of many semantic relations. The two foundational graph representations are established in VectorNet \cite{Gao2020VectorNetEH} and LaneGCN \cite{lanegcn}. 

\subsubsection{Transformer}

Transformers have experienced a huge boost in popularity recently. Their breakthrough in Large Language Models (LLMs) demonstrated the capabilities of this architecture \cite{gpt3,touvron2023llama}. Many recent motion forecasting approaches rely on Transformers \cite{shi2023mtr,shi2022mtra,Nayakanti2022WayformerMF,shi2022motion}. Nevertheless, they have many similarities with GNNs, since they also rely on a graph-based representation. The scene graph is based on the VectorNet \cite{Gao2020VectorNetEH} representation for many approaches. One of the main advantages of Transformers is that their attention mechanism can filter for relevant relations. This enables them to excel on fully connected scene graphs. The downside is that they require lots of data and compute power to work properly. 

Our encoder network relies on a GNN architecture. More specifically, a Heterogeneous Graph Attention network (HGAT). We chose a GNN because it can leverage the diverse relations between different parts of the scene and does not have to learn such distinctions. 

\subsection{Decoder}

Popular decoder architectures have multiple aspects in which they can be differentiated.

\subsubsection{Direct Regression vs Goal Oriented}

One aspect of decoders is whether they directly regress multiple trajectories or if they generate multiple goal points in the first step and then generate trajectories towards these goals in the second step. Both approaches have their pros and cons. Direct regression can accommodate a broad spectrum of agent behaviors \cite{lanegcn}. However, it suffers from slow convergence. Goal-oriented approaches rely on a dense selection of goal candidates \cite{Zhao2020TNTTT}. This reduces the optimization workload. The drawback is that more goal candidates are needed and, therefore, more computational power. Our HGAT relies on direct regression. The refinement module takes entire trajectories as input. These could be considered goal candidates. However, we do not require more goal candidates than direct regression approaches. 

\subsubsection{Discrete Coordinates vs Probability Distribution}

Another aspect is the representation of the predicted trajectories. Some approaches rely on discrete coordinates for the trajectories \cite{lanegcn}. In contrast, others rely on a probability distribution for the trajectories from which they sample discrete trajectories at a later step \cite{multipath, shi2023mtr}. Our refinement module projects the predicted trajectories back into the map. To do so, it requires discrete coordinates. 

\subsubsection{Independent Encoder/Decoder vs Combined Encoder/Decoder}

The most important aspect of the architecture is whether the decoder is an independent part of the network \cite{lanegcn, Cui2022GoRelaGR} or if there is less of a distinction between the encoder and decoder of the network \cite{shi2023mtr, zhou2023query}. Architectures with a clear differentiation between encoder and decoder offer a higher degree of flexibility. The encoder provides a general encoding of the scene. Different decoders can then use that encoding to complete various tasks. This can be motion forecasting but can include various other tasks, such as parked car classification. A drawback of this approach is that the entire scene has to be compressed into a single feature vector. This can cause a loss of information, which causes trajectories to deviate from the available lanes. An architecture without this clear distinction between encoder and decoder can achieve more consistent predictions. But it sacrifices the flexibility a modular architecture offers. Our refinement module aims to provide the modular approach's flexibility with the combined approach's improved consistency.

\section{Scene Graph}\label{scene_graph}

\begin{figure}
    \vspace{1.5mm}
    \centering
    \includegraphics[width=\linewidth]{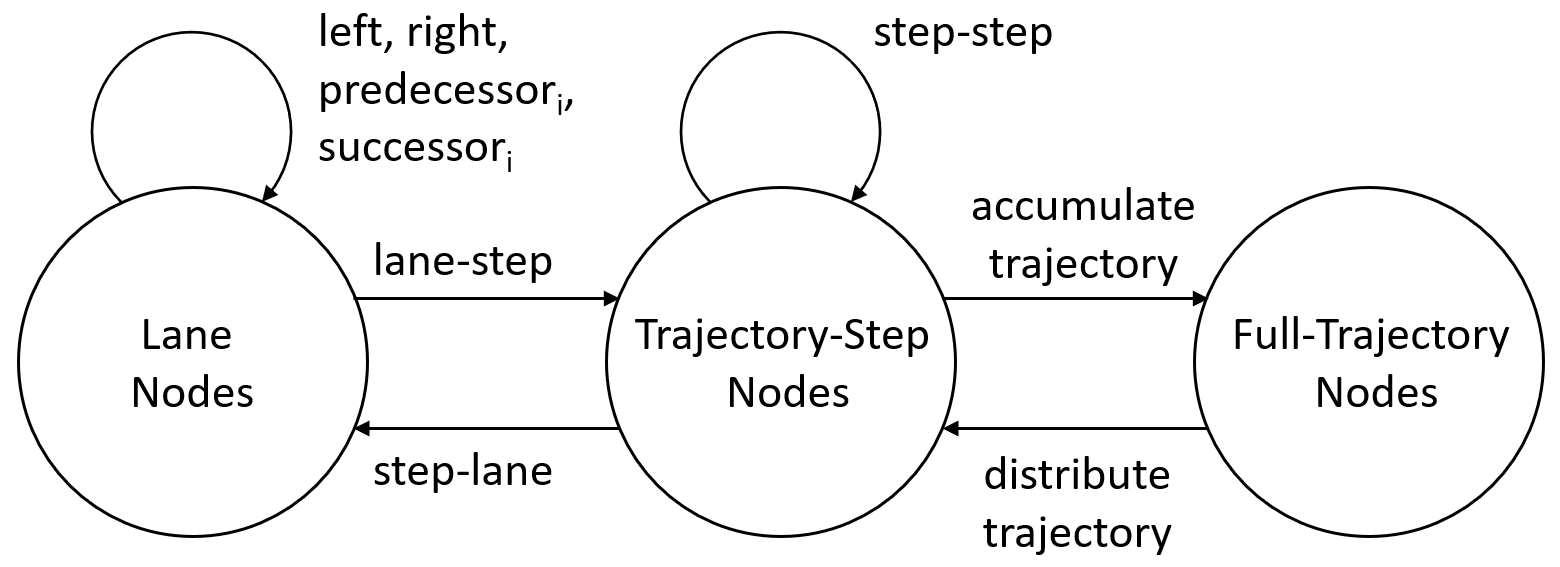}
    \caption{Overview of our heterogeneous graph. There are three types of nodes. The lane nodes represent the map. The trajectory-step nodes represent positional information of trajectories at a specific point in time. The full-trajectory nodes accumulate the information of the corresponding trajectory-step nodes and distribute the combined information back. Lane nodes interact with other lane nodes through the left, right, predecessor and successor edges. Information between lane nodes and trajectory-step nodes is exchanged through the lane-step and step-lane edges. Information between two agents at a specific timestep is exchanged via the step-step edges.}
    \label{fig:graph_relations}
    \vspace{-1mm}
\end{figure}

Our approach relies on a heterogeneous graph that encompasses both the HD map and agent positions. Fig. \ref{fig:graph_relations} depicts the node types and their relations. A visualization of how such a graph looks can be found in Fig. \ref{fig:relations}.

\subsubsection{Node Types}

In the presented heterogeneous graph, we distinguish three distinct node types: Lane nodes, Trajectory-step nodes, and Full-trajectory nodes. Each node type encodes different aspects of pertinent information, thereby contributing to a comprehensive representation of the system under consideration.

\textbf{Lane nodes} constitute a consistent distribution along each mapped lane. They capture specific characteristics of their respective lanes, including positional information, the distance to lane markings, and the categorization of the lane marking type. The formulation of these nodes offers an in-depth understanding of the lane layout within the HD map, providing an integral spatial context for the system.

\textbf{Trajectory-step nodes} represent a particular timestamp for an agent. In the scene graph, they represent the observed timesteps of each agent. They capture dynamic parameters such as the agent's position, velocity, and orientation at a specific time. This time-sensitive information allows for the tracking of an agent's movement and behavior over time, supporting the analysis of temporal patterns and dynamics. In the refinement graph, they represent the positions of the predicted trajectories.

\textbf{Full-trajectory nodes} serve as a compendium of information regarding an agent's activity over all past or future time steps. By aggregating the temporal data encapsulated in the trajectory-step nodes, full-trajectory nodes provide a consolidated view of an agent's trajectory, enabling a comprehensive understanding of its motion pattern over time.

The combination of these three node types creates a robust and multi-faceted representation within the heterogeneous graph. This configuration can then be used for GNN-based motion forecasting.

\subsubsection{Edge Types}

The topology of the heterogeneous graph is characterized by four distinct categories of relationships that interconnect the different node types. These relationships are designed to capture the nuanced interactions within the system and convey the information across different nodes effectively. Not all edges are used in every part of the GNN. 

The first category of relationships pertains to the interconnections between lane nodes. These relationships illustrate the adjacency and continuity of lanes, incorporating the left and right neighbors of a lane node, as well as their predecessors and successors. Through these relationships, the graph can effectively represent the intricate connections and potential paths among lanes. This part of the graph is similar to the lane graph of the LaneGCN architecture.

The second category encompasses the relationships between trajectory-step nodes and lane nodes, which serve as conduits for environmental information transfer, specifically, from the environment to the agents and conversely. To maintain a manageable complexity and avoid information overload, these relationships are constrained to the five nearest agents for a given lane node or the five nearest lane nodes for a given trajectory-step node within a radius of seven meters. This approach ensures that each agent is aware of its immediate surroundings, thereby enhancing its situational understanding and decision-making process.

The third category includes relationships between trajectory-step nodes. These relationships capture the interactions between different agents, taking into consideration the five nearest agents within a larger radius of 100 meters. Such a configuration allows the system to account for more distant, yet potentially influential, agents that might affect the decision-making of the focal agent.

The fourth and final category involves the relationships between trajectory-step nodes and their corresponding full-trajectory nodes. These relationships relay the time-specific state of a trajectory-step node to its respective full-trajectory node. This design enables the system to compile a comprehensive history of each agent's motion over time, providing a temporal context.

These relationship categories form the backbone of the heterogeneous graph structure, enabling the complex interplay between spatial and temporal elements within the GNN.

\begin{figure*}
\vspace{1.5mm}
\centering  
\begin{subfigure}{0.5\textwidth}  
  \centering  
  \includegraphics[width=0.9\linewidth]{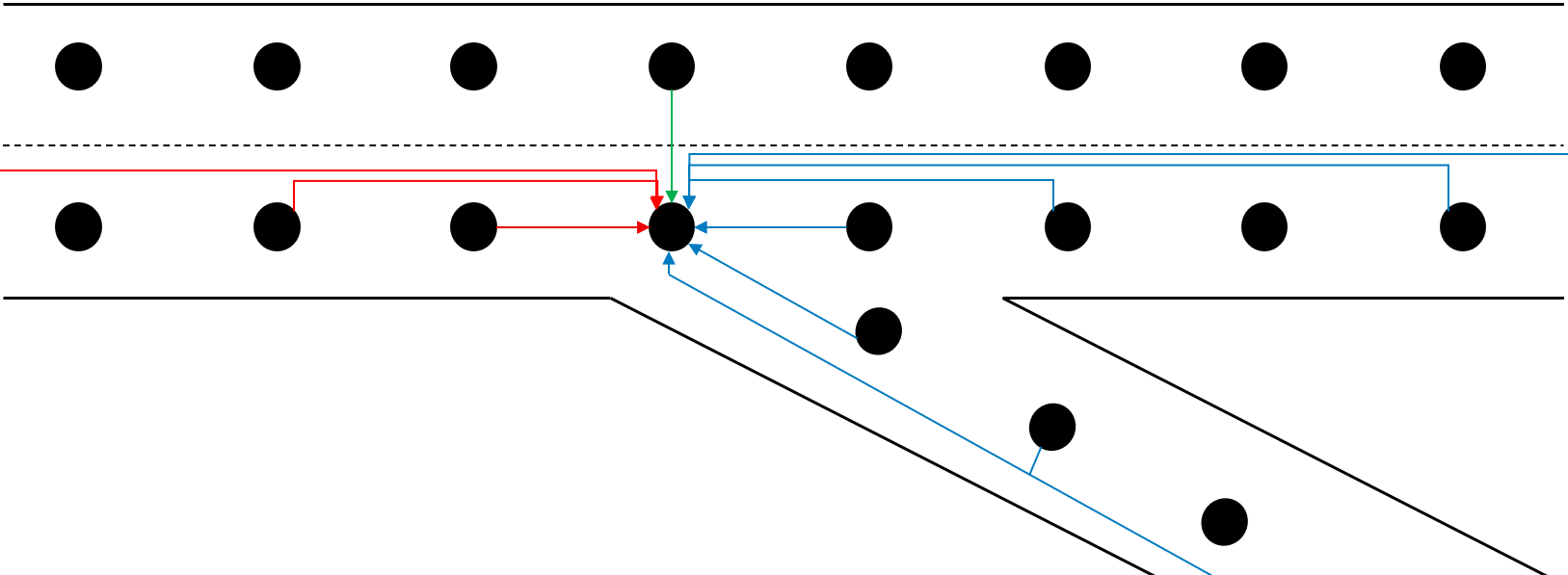}  
  \caption{Lane / Lane interaction}  
  \label{fig:l2l}  
\end{subfigure}%
\begin{subfigure}{0.5\textwidth}  
  \centering  
  \includegraphics[width=0.9\linewidth]{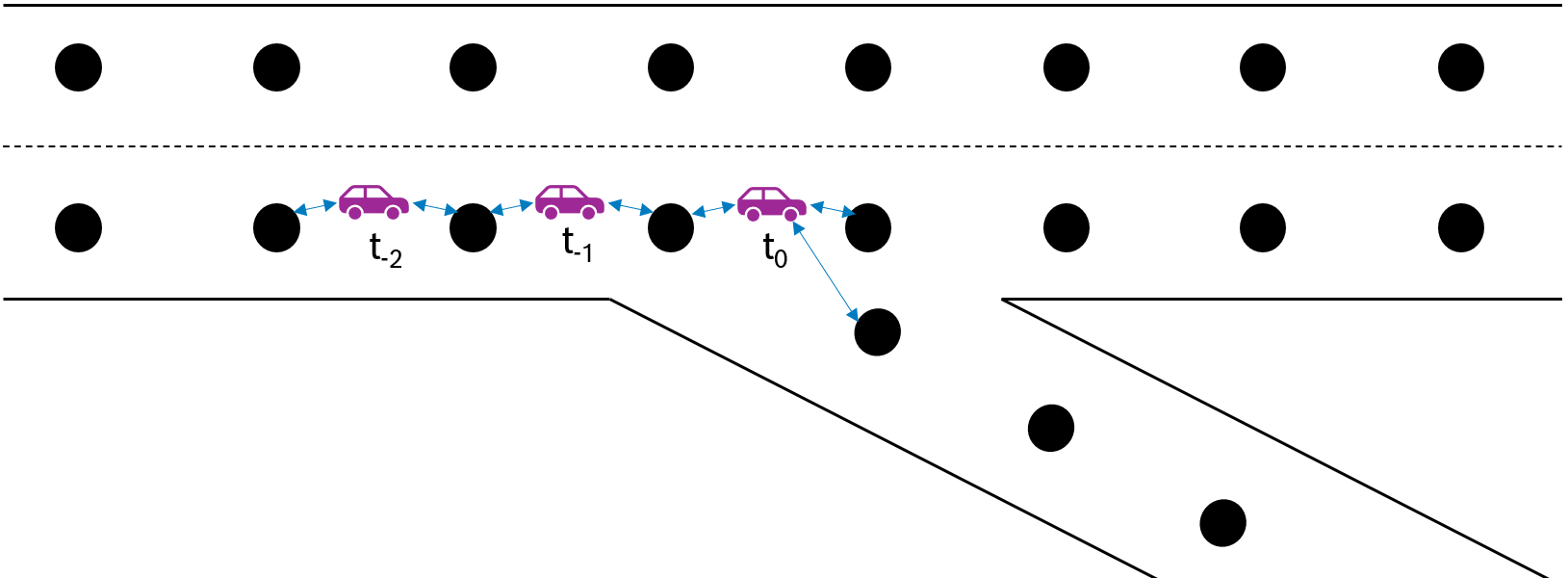}  
  \caption{Lane / Trajectory-step interaction}  
  \label{fig:l2a}  
\end{subfigure}  
\newline  
\begin{subfigure}{0.5\textwidth}  
  \centering  
  \includegraphics[width=0.9\linewidth]{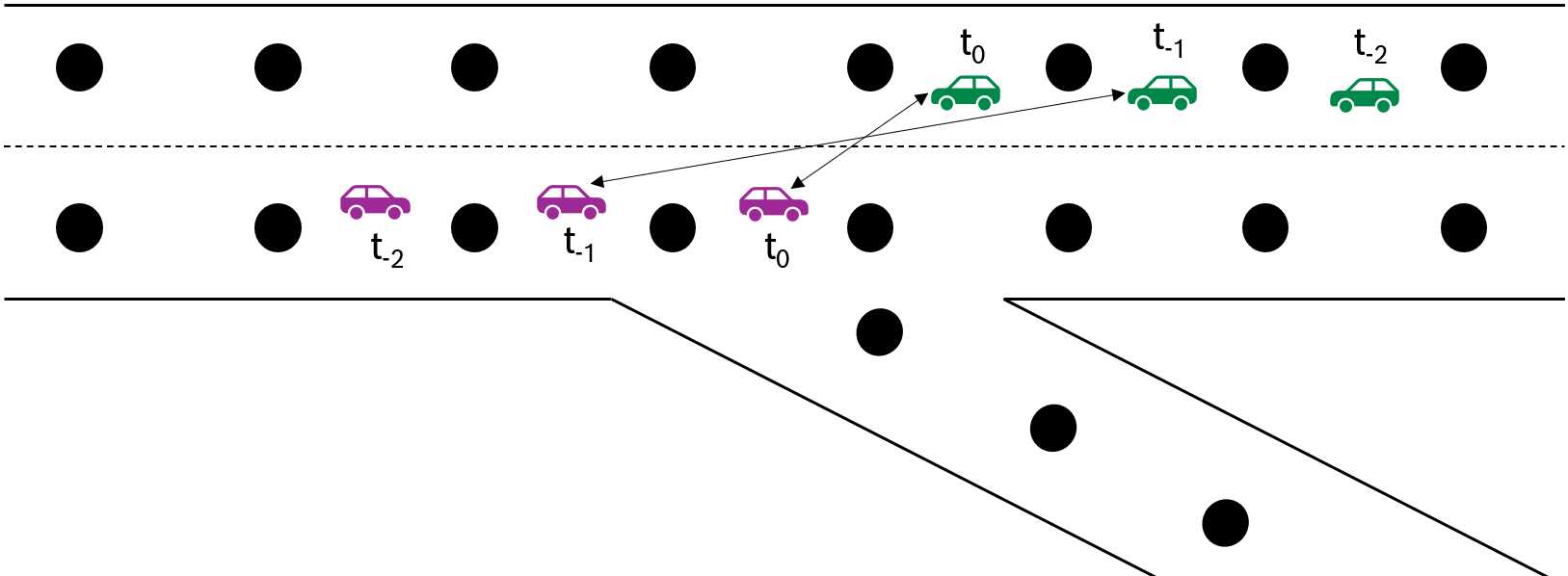}  
  \caption{Trajectory-step / Trajectory-step interaction}  
  \label{fig:a2a}  
\end{subfigure}%
\begin{subfigure}{0.5\textwidth}  
  \centering  
  \includegraphics[width=0.9\linewidth]{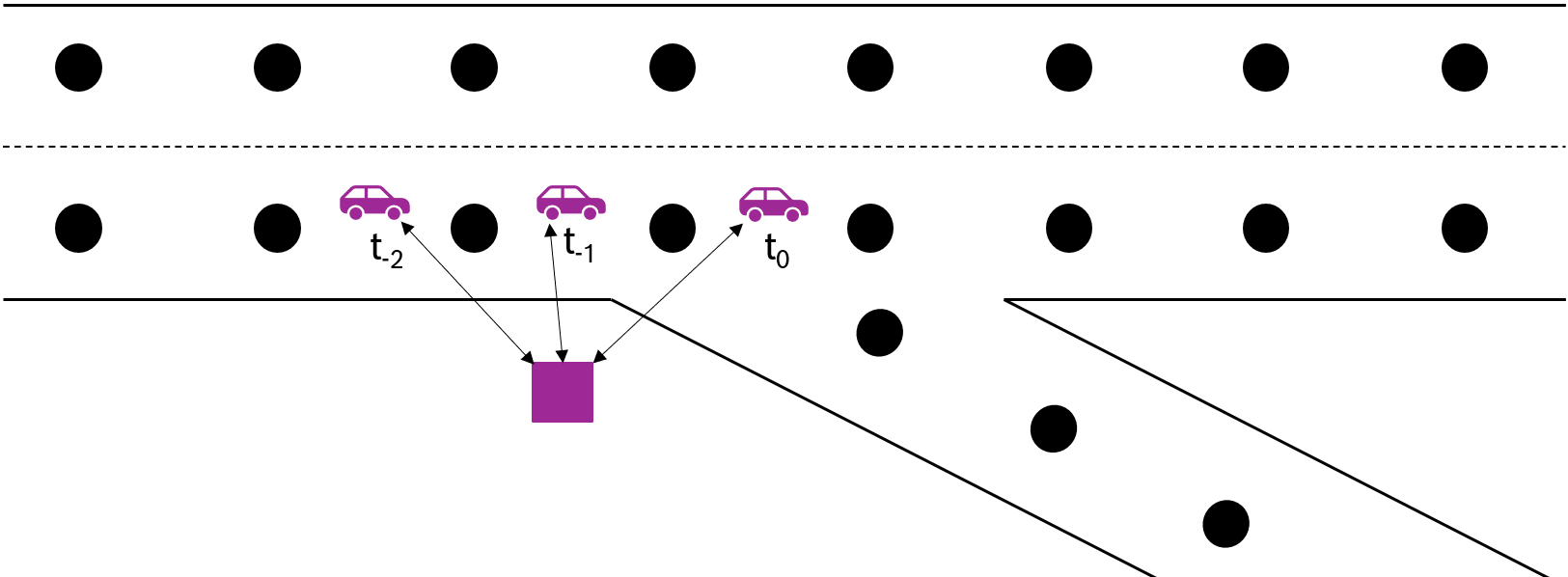}  
  \caption{Trajectory accumulation / distribution}  
  \label{fig:sub4}  
\end{subfigure}  
\caption{Overview of how the heterogeneous graph structure looks like in an example scene. (a) visualizes the interaction between lane nodes. Green arrow: left neighbor, red arrows: predecessor nodes, blue arrows: successor nodes. (b) demonstrates the connections between trajectory-step nodes (depicted as cars) and lane nodes. Trajectory-step nodes are connected to nearby lane nodes with a similar orientation. (c) shows the interaction between different trajectory steps. Only trajectory steps of the same point in time within a certain distance threshold can interact with each other. $t_i$ indicates the timestep and the color of the node indicates different agents. (d) each observed trajectory has its own full-trajectory node. This node accumulates the information of the corresponding trajectory-step nodes and disperses the combined information back to the individual trajectory steps.}  
\label{fig:relations}  
\vspace{-1mm}
\end{figure*}

\section{Heterogeneous Graph Attention Network}
We propose a custom variant of a Heterogeneous Graph Attention network (HGAT). This GNN supports heterogeneous node and edge types, edges with additional feature vectors, and an attention mechanism as proposed by GATv2 \cite{gatv2}. It uses an attention function in which the nonlinear activation $\sigma$ separates the learned attention parameters $a$ from the processed feature vector $f_{j,r,i}^h$. This is done to prevent the problem with static attention of the original GAT \cite{gat}.

\begin{equation}
    \alpha_{j,r,i}^h = \textrm{softmax}_i \left( (a^h)^\top \cdot \sigma \left(  f_{j,r,i}^h \right) \right)
\end{equation}

In these functions, $(j,r,i)$ denotes the edge between the source node $j$ and the target node $i$ with a relation type $r$. Our HGAT supports multiple attention heads $h$. Depending on the relation type $r$, the target node features $\textbf{v}_i$, neighboring node features $\textbf{v}_j$ and edge features $\textbf{e}_{j,r,i}$ are all encoded using the fully connected layers ($W_{s,r}^h, W_{n,r}^h, W_{e,r}^h$), where $\parallel$ denotes the concatenation operation. The feature vector $f_{j,r,i}^h$ used in the attention is defined as:

\begin{equation}\label{eq:att_feat}
    f_{j,r,i}^h = W_{s,r}^h \textbf{v}_i \parallel W_{n,r}^h \textbf{v}_j \parallel W_{e,r}^h \textbf{e}_{j,r,i}.
\end{equation}

The node update $\textbf{v}_i'$ consists of a learned residual $W_{res, t}$ of the previous node features $\textbf{v}_i$ which is also dependent on the node type $t$ and the accumulated neighboring node features weighted by the attention score. This accumulated and weighted feature vector is denoted as $\hat{f}_i$

\begin{equation}
    \textbf{v}_i' = \sigma \left( W_{res, t} \textbf{v}_i + \hat{f}_i \right) 
\end{equation}

The accumulation of the neighboring node features is defined in Equation \ref{eq:f_i}. The neighborhood of node $i$ is given by $\mathcal{N}(i)$. The encoded features of the neighboring node $\textbf{v}_i$ and the encoded edge features $\textbf{e}_{j,r,i}$ are weighted by the attention score $\alpha_{j,r,i}^h$. This is done for all $H$ attention heads, which are then concatenated.

\begin{equation}\label{eq:f_i}
    \hat{f}_{i} = \sum_{(j,r) \in \mathcal{N}(i)} \parallel_{h=1}^H \alpha_{j,r,i}^h \left( W_{n,r}^h \textbf{v}_j + W_{e,r}^h \textbf{e}_{j,r,i} \right)
\end{equation}

\section{NETWORK ARCHITECTURE}

We propose a custom GNN architecture to improve the well-established LaneGCN architecture in two significant ways. First, the entire scene encoder is unified in a single Heterogeneous Graph Attention network. Additionally, we introduce a separate refinement module that projects the predicted trajectories back onto the HD map. This refinement module can be easily adapted to many other motion forecasting approaches. The following encoders compute the feature vectors of each corresponding node for the GNN and then use these feature vectors to encode the map and the entire scene further.

\begin{figure}
    \vspace{1.5mm}
    \centering
    \includegraphics[width=1\linewidth]{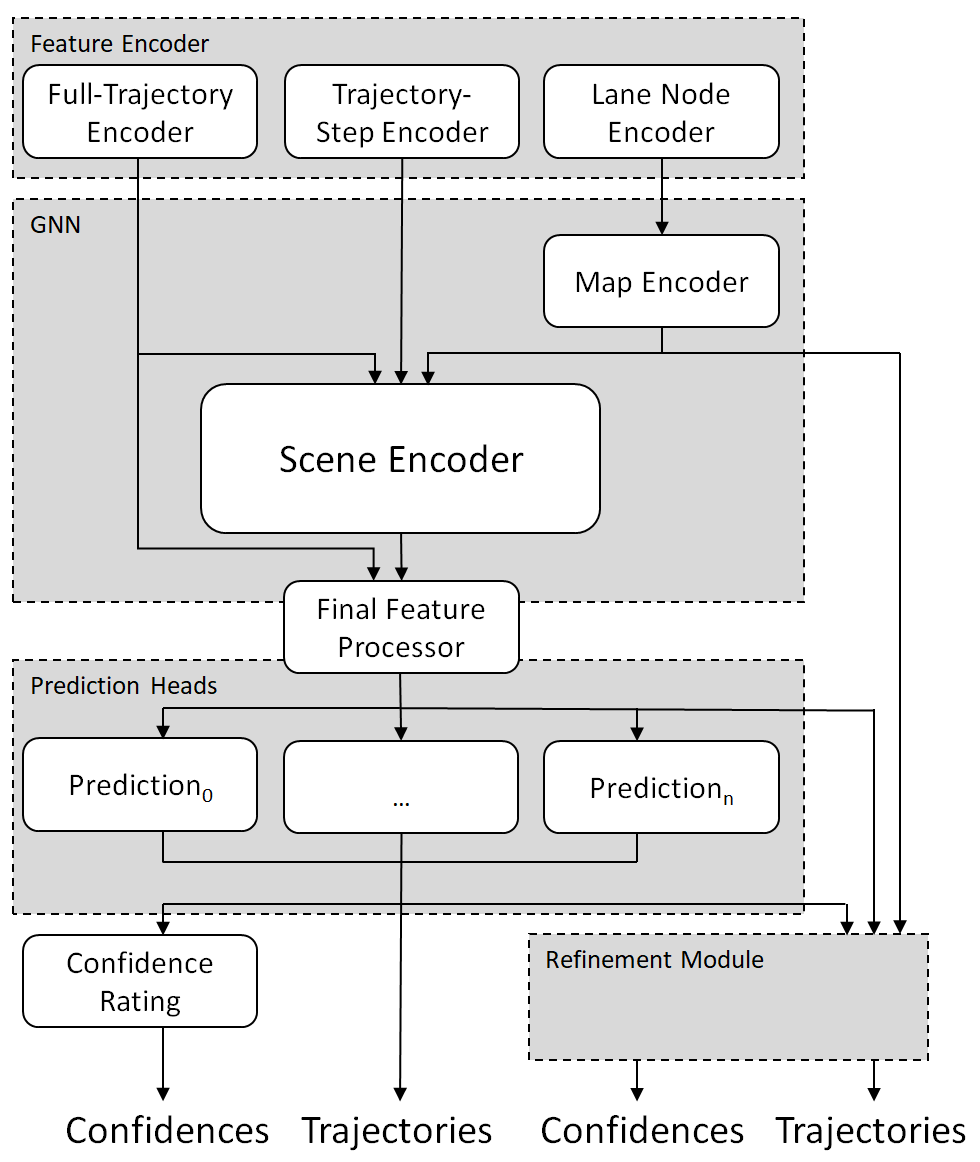}
    \caption{The network architecture starts by encoding varying-sized input features into uniform-sized feature vectors. Map features are encoded with a GNN, followed by encoding the whole scene. Trajectory node features are extracted and merged with residual trajectory features for final feature encoding. These final features are processed by independent prediction heads to create trajectory proposals. If no refinement module is employed, the confidence module ranks the predicted trajectories. If a refinement module is used, it takes in trajectory proposals, final feature vectors and encoded lane nodes to predict and rank the final trajectories.}
    \label{fig:architecture}
    \vspace{-5mm}
\end{figure}

\subsection{Encoder}

\subsubsection{Full-Trajectory Encoder}
The Full-Trajectory Encoder encodes the entire trajectories of various agents over time. This encoder captures the sequential nature of agents' movements, thereby modeling the temporal dynamics of agent trajectories. It is designed to transform raw trajectory data into a high-dimensional feature space where the encoded features can effectively represent the complex motion patterns of the agents. It consists of a residual convolutional network with two layers and batch normalization.

\subsubsection{Trajectory-Step Encoder}
The Trajectory-Step Encoder is designed to encode the states of agents at a specific timestep. It captures the instantaneous state of an agent, including position, velocity, acceleration, and other relevant state variables. By encoding these state variables, it provides a snapshot of the agent's current situation, which complements the temporal information provided by the Full-Trajectory Encoder. It consists of four linear layers with batch normalization and a residual connection.

\subsubsection{Lane Node Encoder}
The Lane Node Encoder is tasked with encoding the features of all nodes in the HD map. By encoding the features of these map nodes, the Lane Node Encoder can capture the static environmental context in which the agents operate. It has the same architecture as the Trajectory-step Encoder.

\subsubsection{Map Encoder}
The map encoder consists of four layers of our HGAT and encodes only the lane nodes with the lane/lane interaction edges in our scene graph (Fig. \ref{fig:l2l}). This enriches the lane nodes with information from other reachable lane nodes.

\subsubsection{Scene Encoder}
The scene encoder then passes the entire scene graph with all lane nodes, trajectory-step nodes and full-trajectory nodes through four layers of our HGAT. In the end, the full-trajectory nodes contain all the relevant information for each corresponding trajectory-step and their surroundings. 

\subsubsection{Final Feature Processor}
The final feature processor combines the residual features from the full-trajectory encoder and the features from the full-trajectory nodes after they have been processed by the scene encoder. It consists of three linear layers with batch normalization.

\subsection{Motion Forecaster}

\subsubsection{Trajectory Prediction Heads}
There are $n$ trajectory prediction heads for each type of agent. They independently predict trajectories for each agent. A prediction head consists of seven linear layers with batch normalization. Besides the trajectories, these heads also generate a feature vector with additional information which is then used in the confidence head or in the refinement module.

\subsubsection{Confidence Rating}
The confidence rating head ranks the different trajectories for each agent. There is a separate confidence head for each type of agent. It consists of five linear layers with batch normalization. The output is passed through a softmax function to get a probability distribution.

\subsection{Refinement Module}
The refinement module takes previously generated trajectories and map information to improve the predicted trajectories. This is done by generating a new graph based on the predicted trajectories and the previously generated lane graph. This graph follows the same structure as the previously used scene graph described in Section \ref{scene_graph}. Each trajectory-step node now represents a point of one predicted trajectory. The full-trajectory node now accumulates the features of a single predicted trajectory. While the node types are the same as in the previous scene graph, the edges will be different. The edges between lane nodes and the edges between trajectory-step nodes are not used. The edges between lane nodes and trajectory-steps will be dynamically generated.

\subsubsection{Requirements}
This refinement module is not limited to the previously introduced scene encoder and prediction architecture. It can be easily adapted to many other architectures as long as the following requirements are satisfied. 
\begin{itemize}
    \item \textbf{Lane nodes} need to have a coordinate key and location-specific features.
    \item \textbf{Trajectory-step nodes} need to have a coordinate key, too. They also need an initial feature vector with some information about the trajectory.
    \item \textbf{Full-trajectory nodes} only need an initial feature vector with additional information about the scene.
\end{itemize}

\subsubsection{Integration into our Network}
The output of our map encoder already fulfills the requirements for the lane nodes, which is why we reuse these lane nodes in our refinement module. Our prediction headers provide the trajectories, which are transformed into their respective trajectory-step nodes. They also provide additional features which are used as initial feature vectors. Lastly, a full-trajectory node is generated for each predicted trajectory and initialized with the final feature vector used for the prediction of the corresponding agent.

\subsubsection{Information Flow}
\begin{figure}
    \vspace{1.5mm}
    \centering
    \includegraphics[width=\linewidth]{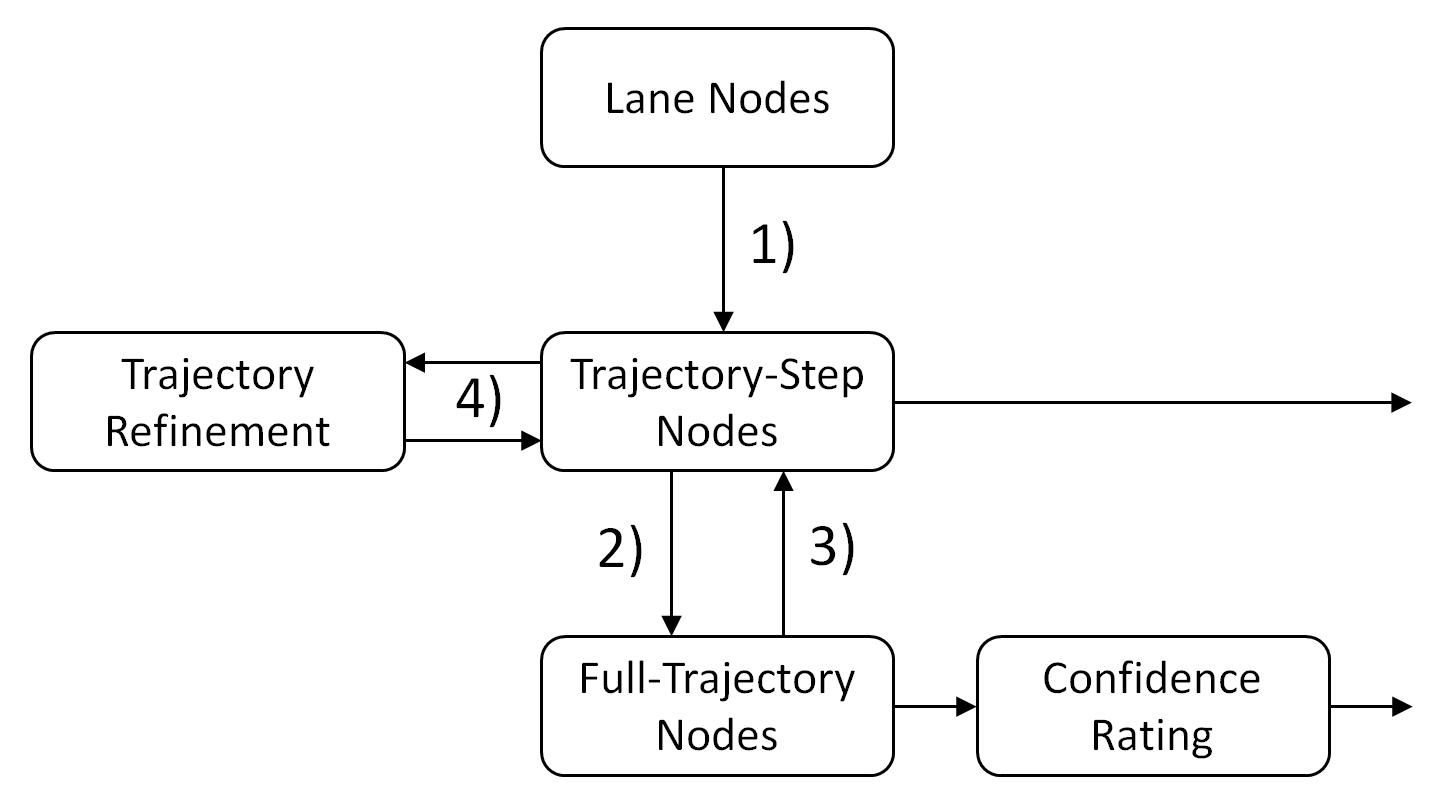}
    \caption{Overview of the refinement module. It leverages parts of the same heterogeneous graph. 1) Map information is transported to the predicted trajectory steps. 2) Corresponding trajectory steps are accumulated in the full-trajectory nodes. 3) The accumulated information is distributed back to the individual trajectory-step nodes. 4) The refinement network modifies the current trajectory prediction. In the end, the trajectories are rated in the confidence network.}
    \label{fig:refinement_module}
    \vspace{-1mm}
\end{figure}

Our process iterates through a series of steps depicted in Fig. \ref{fig:refinement_module}.

\textbf{Step 1)} The current trajectory-step coordinates are compared to the lane node coordinates. The comparison results in the generation of dynamic edges in the graph, with the five closest lane nodes connected to each trajectory step node.
These edges are then used in a transformer convolution layer~\cite{transconv} to enrich the trajectory-step nodes with the lane node features. This step allows the trajectory nodes to assimilate crucial information from the surrounding lane nodes, enhancing their contextual understanding.

\textbf{Step 2)} In the subsequent phase, another transformer convolution layer is utilized to accumulate the information of the trajectory-steps in the full-trajectory nodes.

\textbf{Step 3)} Post accumulation, the compiled information is disseminated back to the trajectory step nodes via yet another transformer convolution layer. 

\textbf{Step 4)} In the final stage, the trajectory refinement network computes an offset to the original trajectory. This offset is then applied to the trajectory nodes, producing a refined trajectory that is more aligned with the actual lane nodes.

This completes one iteration, and the subsequent iteration can start with newly generated dynamic edges. After all iterations are completed, the information of the trajectory step nodes is accumulated one last time in the trajectory nodes and passed to the confidence network to assign the correct confidences to the predicted trajectories.

\subsection{Learning}
In order to train the network, we deploy the loss function that has been established in LaneGCN \cite{lanegcn}. It consists of a weighted loss function in which a regression loss is used for the trajectory and a max-margin loss is used for the confidence. All the modules are differentiable and can be trained in an end-to-end way.

\begin{equation}
L(\text{agent}) = L_{\text{conf}} + \alpha L_{\text{traj}},
\end{equation}
where $\alpha=1.0$. 

Even though our focus is only on predicting the trajectory of a single agent, we found it advantageous to include the predictions of all agents $\mathcal{A}$ in the loss function. The trajectories of other agents were assigned a lower weight based on their importance $w(a)$.

\begin{equation}
L_{total} = \sum_{a\in\mathcal{A}} w(a)L(a)
\end{equation}

\section{Experiments}

\subsection{Dataset}

We decided to use the Argoverse 2 Motion Forecasting dataset \cite{Argoverse2} for our experiments. It is a well-established benchmark for trajectory prediction and a successor to the already successful first iteration of the dataset \cite{Argoverse}. The dataset is publicly available. It was recorded in multiple US cities and contains detailed HD maps and over 250k traffic scenarios with a duration of 11s each. 200k scenarios are used for training, 25k are used for validation and the remaining 25k are used for testing. The test data does not include ground truth labels and needs to be evaluated by the online benchmark. The individual 11s scenarios are split into two parts. The first 5s are used as input for the prediction and the remaining 6s are used as ground truth. Each scenario contains a focal track, which is the main focus of the single agent prediction benchmark. Additionally, there are also the trajectories of other agents, which are categorized into scored tracks, unscored tracks and track fragments. The agents are also classified into different agent types (i.e., vehicle, pedestrian, bus, cyclist, motorcyclist).  

\subsection{Metrics}
The Argoverse 2 Motion Forecasting dataset is evaluated using three primary metrics, namely, Average Displacement Error (ADE), Final Displacement Error (FDE), and Miss Rate (MR). ADE describes the average Euclidean distance between the prediction and ground truth at each timestep. FDE only measures the Euclidean displacement at the trajectory endpoint. A trajectory is considered a miss if the FDE for a given trajectory is larger than 2m. For the benchmark, a maximum of six trajectory proposals is accepted. This is considered by using the minimum FDE (minFDE) and minimum ADE (minADE) for the top K proposals ranked by confidence. The evaluation is done with K=1 and K=6. Finally, the brier-minFDE combines the confidence $c$ in the best prediction with the minFDE. This is done by adding $(1-c)^2$ to the minFDE. The official ranking is determined by the brier-minFDE.

\subsection{Baseline}
In selecting a baseline for our experiments, we opted for LaneGCN for various reasons. LaneGCN is considered a highly influential architecture, providing a foundational structure upon which numerous subsequent approaches have been developed and refined \cite{gilles2022gohome,gilles2021thomas, Cui2022GoRelaGR}. In our research, we propose a novel approach that condenses the intricate multistep architecture of LaneGCN into a singular GNN, streamlining the process while preserving the robustness of the original structure. Furthermore, our research is designed to demonstrate that our versatile refinement module significantly improves the foundational LaneGCN architecture. By extension, given the influence of LaneGCN on later models, it follows that our refinement module will have a positive effect on these derivative works as well, potentially leading to broader applications and advancements in the field.

\section{RESULTS}
First we will evaluate our approach on the standard Argoverse 2 benchmark metrics and compare them to our baseline. Then, we will show some qualitative results that demonstrate the effects of our refinement module. In the end, we will demonstrate the insights we can obtain by analyzing the attention in our network.

\subsection{Metrics}

The Argoverse 2 benchmark metrics can be seen in Table~\ref{tab:test}. Our HGAT model without the refinement module is better in all metrics. When adding the refinement module to the already trained model, we see notable improvements in the K=6 metrics, whereas the effects on the K=1 metrics are smaller. The biggest effects of our refinement module can be seen when training it in an end-to-end fashion. Here, we observe significant performance gains across all metrics.

\begin{table*}
    \vspace{1.5mm}
	\begin{center}
		\caption{Results on Argoverse 2 Motion Forecasting benchmark (test set)} \label{tab:test}
		\begin{tabular}{l||ccc|cccc}
			\hline
			\multirow{2}{*}{Model} & \multicolumn{3}{c|}{K=1} & \multicolumn{4}{c}{K=6}\\
			& minADE & minFDE & MR & minADE & minFDE & MR & \textbf{brier-minFDE}\\
			\hline
			LaneGCN \cite{lanegcn}      
                & 2.43 & 6.51 & 0.71 & 0.91 & 1.96 & 0.30 & 2.64\\
            \hline
            Our Model (w/o refinement)          
                & 2.14 & 5.53 & 0.69 & 0.90 & 1.84 & 0.29 & 2.52\\ 
			Our Model (w/ refinement + pretrained) 
                & 2.19 & 5.50 & 0.68 & 0.86 & 1.71 & 0.24 & 2.39\\
            Our Model (w/ refinement e2e)
                & \textbf{2.01} & \textbf{5.13} & \textbf{0.66} & \textbf{0.81} & \textbf{1.60} & \textbf{0.21} & \textbf{2.23}\\
			\hline
		\end{tabular}
	\end{center}
    \vspace{-5mm}
\end{table*}

\subsection{Effects of Refinement}

\begin{figure} 
\vspace{1.5mm}
\centering  
\begin{subfigure}{0.25\textwidth}  
  \centering  
  \includegraphics[width=0.75\linewidth]{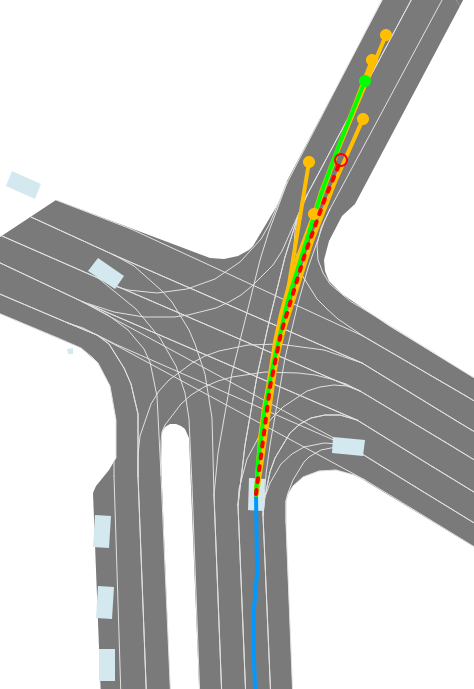}  
  \caption{w/o refinement}  
  \label{fig:s14_wo_ref}  
\end{subfigure}%
\begin{subfigure}{0.25\textwidth}  
  \centering  
  \includegraphics[width=0.75\linewidth]{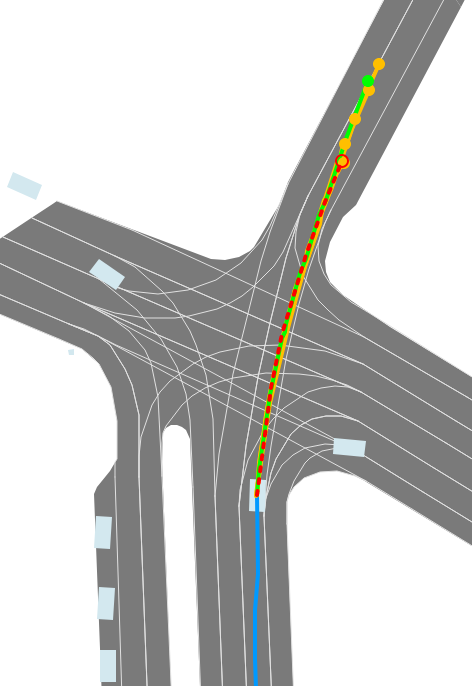}  
  \caption{w/ refinement}  
  \label{fig:s14_e2e} 
\end{subfigure}  
\caption{(a) Without the refinement module, the exact lane boundaries are lost and the prediction drifts into the other lane. (b) With the refinement module, all the map information can be used to shift the trajectories toward the actual lane. Blue line: observed history, red line: ground truth future trajectory, green line: predicted trajectory with highest confidence, orange lines: other predictions.}   
\label{fig:demo} 
\vspace{-1mm}
\end{figure}

The problem with approaches that solely rely on a single feature vector to encode a scene into a latent space is that they tend to miss important details. The general road layout is conserved. But the exact curvature of lanes might get lost. This can be seen in Fig. \ref{fig:s14_wo_ref}, which shows the prediction of our HGAT without the refinement module. The scene shows an intersection where the road curves slightly to the right. For the most part, the predictions follow this curvature. However, multiple trajectories drift into the neighboring lane on the left side. When adding our refinement module, this effect is significantly reduced, as can be seen in Fig. \ref{fig:s14_e2e}. Here, all predicted trajectories follow the lane curvature. 

\subsection{Attention Insights}

\begin{figure}[!ht]
\centering  
\begin{subfigure}{0.20\textwidth}  
  \centering  
  \includegraphics[width=1\linewidth]{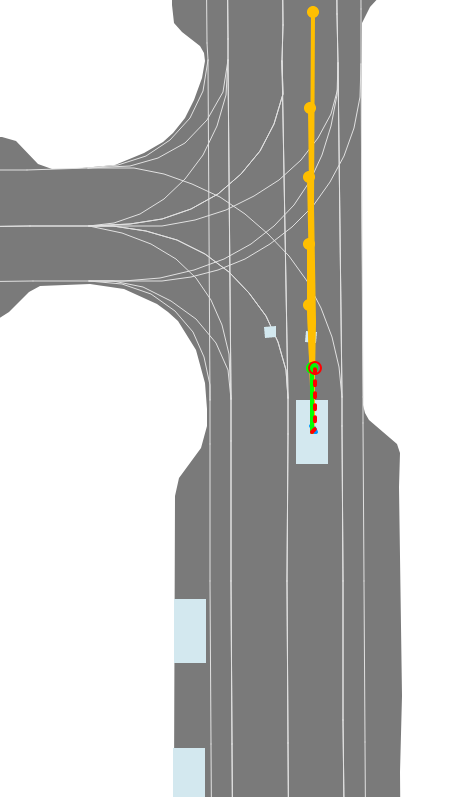}  
  \caption{Prediction}  
  \label{fig:s107_pred}  
\end{subfigure}%
\begin{subfigure}{0.3\textwidth}  
  \centering  
  \includegraphics[width=0.7\linewidth]{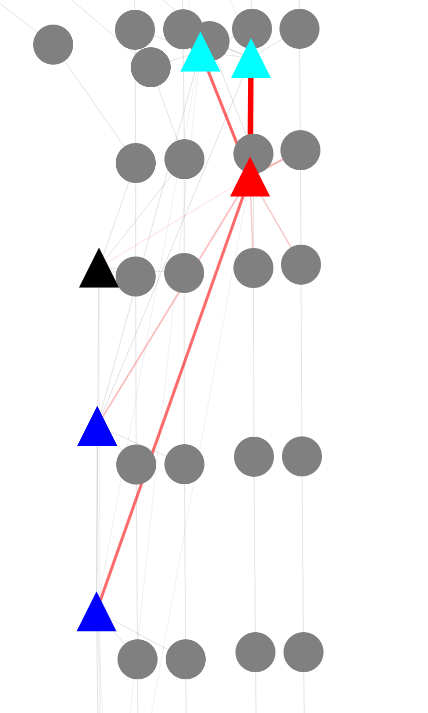}  
  \caption{Attention Graph}  
  \label{fig:s107_att}  
\end{subfigure}  
\caption{The scene shows a car waiting for two pedestrians to cross from the right to the left. (a) shows the predictions. The one with the highest confidence (green) mirrors the ground truth (red), which shows a slow acceleration after the pedestrians clear the path. (b) When analyzing the attention score inside the network, we can see that the network pays a lot of attention to the pedestrians, especially to the slower one on the right. This indicates that the pedestrian is considered in our prediction. The attention scores are visualized with red edges. The broader and redder an edge, the higher the attention score for this connection. The triangles are agents. Red: focal agent, light blue: pedestrian, dark blue: car, black: static obstacle. The gray circles are lane nodes.}  
\label{fig:demo_attention}  
\vspace{-5mm}
\end{figure} 

Our unified HGAT architecture provides unique insight into the network through the different attention scores in our heterogeneous scene graph. Fig. \ref{fig:demo_attention} shows a scene in which a car is waiting for pedestrians to cross the street from right to left. The predicted trajectories can be seen in Fig. \ref{fig:s107_pred}. The trajectory with the highest confidence (green) matches the ground truth (red) well. This prediction models a trajectory in which the car waits for the pedestrians to cross the street further and then slowly accelerates. The other less likely trajectories have a more aggressive acceleration profile. When looking at the attention scores inside our HGAT network, we can verify that the car is paying attention to the pedestrians. The right pedestrian is the most crucial in the scene since he is currently directly in the car's path and will clear it last. This pedestrian also has the highest attention score, indicating that the pedestrian is considered in our prediction.

\subsection{Ablation Study}

We performed an ablation study to evaluate the impact of different edge types in our heterogeneous graph. The results on the official test set can be seen in Table \ref{tab:abl}. We removed various edge types from the heterogeneous graph (Fig. \ref{fig:graph_relations}) and retrained the network with the refinement module. The graph used in the refinement module was not modified.

\begin{table}[H]
    \vspace{1.5mm}
	\begin{center}
		\caption{Ablation study on impact of different edge types} \label{tab:abl}
		\begin{tabular}{cccc||c}
			\hline
			\multicolumn{4}{c||}{Removed edge types} & \multicolumn{1}{c}{K=6}\\
            lane-step & step-lane & step-step & distribute traj. & brier-minFDE\\
			\hline
			X & X & X & X & 2.39\\
              & X & X & X & 2.31\\
            \hline
            X & X &   &   & 2.30\\
			   & X &   &   & 2.26\\
              &   & X &   & 2.27\\
              &   &   & X & 2.26\\
            \hline
             &  &  &  & \textbf{2.23}\\
			\hline
		\end{tabular}
	\end{center}
\end{table}

\section{CONCLUSION}
We have demonstrated that our heterogeneous graph attention network improves upon the highly influential LaneGCN architecture. Furthermore, we have shown that our refinement module provides a significant improvement to the results and increases the consistency of the prediction in relation to the map. Both the metrics and the observed results show clear advantages. Since there are only a few requirements to apply the refinement module to other architectures, it is quite easy to apply the module to other architectures, especially if the architecture is based on LaneGCN. We have also demonstrated how the attention scores of our unified HGAT network give valuable insights into the network and help in explaining how certain elements in the scene influence the prediction.









{\small
\bibliographystyle{IEEEtran}
\bibliography{sources}
}

\end{document}